\documentclass[english,12pt]{article}
\usepackage[letterpaper, portrait, margin=1.2in]{geometry}
\usepackage[T1]{fontenc}
\usepackage[latin9]{inputenc}
\usepackage{graphicx}
\usepackage{multirow}
\usepackage{caption}
\usepackage{subcaption}
\usepackage[square,sort,comma,numbers]{natbib}
\usepackage{setspace}
\usepackage{morefloats}
\usepackage{makecell}
\usepackage{authblk}
\makeatletter

\providecommand{\tabularnewline}{\\}

\usepackage{color}
\definecolor{rev}{RGB}{0,0,0} 
\usepackage{babel}
\graphicspath {{other_figures/}}

\makeatother

\begin{document}

\title{Early Predictions of Movie Success: the Who, What, and When of Profitability}

\date{}
\author[1]{Michael T. Lash\thanks{michael-lash@uiowa.edu, michaeltlash.com}} 
\author[2]{Kang Zhao\thanks{kang-zhao@uiowa.edu, +1-319-335-3831 (Corresponding author)}}

\affil[1]{Department of Computer Science, The University
of Iowa, 318 MacLean Hall, Iowa City, IA 52242, United States}
\affil[2]{Department of Management Sciences, Tippie College
of Business, The University of Iowa, S224 PBB, Iowa City, IA, 52242, United States}

\maketitle

\begin{abstract}
This paper proposes a decision support system to aid movie investment
decisions at the early stage of movie productions. The system predicts
the success of a movie based on its profitability by leveraging historical
data from various sources. Using social network analysis and text
mining techniques, the system automatically extracts several groups
of features, including ``who'' are on the cast, ``what'' a movie
is about, ``when'' a movie will be released, as well as ``hybrid''
features that match ``who'' with ``what'', and ``when'' with
``what''. Experiment results with movies during an 11-year period showed that the system outperforms benchmark methods by a large margin in predicting movie profitability. Novel features we proposed also made great contributions to the prediction. In addition to designing a decision support system with practical utilities, our analysis of key factors for movie profitability may also have implications for theoretical research on team performance and the success of creative work.
\end{abstract}
\providecommand{\keywords}[1]{\textbf{\textit{Keywords: }} #1} 
\newpage{}

\doublespacing 

\section{Introduction}

The motion picture industry is a multi-billion dollar business. In
2012, the United States and Canada saw a total box office revenues topping
\$10.8 Billion \citep{MPAA2012}. For films are released in U.S. and Canada, the average investment is \$65 million per movie \citep{Mueller2011}. Despite the amount of in-flux capital, the certainty with regard to movie success is
largely uncertain, with ``hits'' and ``flops'' released almost
every year. While researchers have undertaken the task of predicting
movie success using various approaches, they attempted to predict
box office revenues, or theater admissions. However, from an investor's
standpoint, one would want to be as assured as possible that his/her
investment will ultimately lead to returns or profits. For instance,
``Evan Almighty'' earned a high gross revenue of \$100 million,
but cost \$175 million to produce; while ``Super Troopers'' cost
\$3 million, but earned \$18.5 million. The latter is certainly a
better bet for investors. In fact, among movies produced between 2000
and 2010 in the U.S., only 36\% had a box office revenue higher than
its production budget, which further highlights the importance of
making the right investment decisions. Therefore, our work defines
a movie's success as its profitability and attempts to predict such
success in an automated way to better support movie investors' decisions.

In order to support investment decisions on a movie, the prediction
of profitability has to be provided at the very early stage of the
movie's production. Consequently, our prediction of movie success
can only leverage data that is available when a movie is still being
planned. Predictions that are made right before \citep{Eliashberg2000}
or after \citep{Zhang2009,Boccardelli2008,Meiseberg2013} the official
release may have more data to use and get more accurate results, but
they are too late for investors to make any meaningful decision. Building
upon our previous work \citep{Lash2015}, this research proposes a
Movie Investor Assurance System (MIAS) to provide early predictions
of movie profitability. Based on historical data, the system automatically
extracts important characteristics for each movie, including ``who'' will
be involved in the movie, ``what'' the movie is about, ``when''
the movie will be released, and the match between these features.
It then uses various machine learning methods to predict the success
of the movie with different criteria for profitability.

The overarching research question for this paper is to predict movie
profitability using data only available during the pre-production
stage of movie development in order to support investment decisions.
The main contributions of this research are in 3 areas:  \textcolor{rev}{\textit{First}, this decision support system is the first to harness machine learning, text mining, and social network analytics into one comprehensive decision-support system to predict movie profitability (rather than revenue), especially doing so during early stages of movie production, and with minimal human interventions. \textit{Second}, our research proposes several novel features, such as dynamic network features, plot topic distributions, the match between ``what'' and ``who'', the match between ``what'' and ``when'', and the use of profit-based star power measures. We showed that these features all contribute to the system's performance in predicting profitability. \textit{Third}, our system is the first to collect different types of data (including structured data, network data, and unstructured data) from different freely-available sources, and fuse them for predicting the success of movies.}

The remainder of the paper is organized as follows: after reviewing
related research in Section 2, we describe the framework of our system.
Section 4 introduces how we extracted different features for the prediction.
This is followed by the evaluation of our system using historical
data and the analysis of the key factors behind movie success predictions.
The paper concludes with discussions of limitations and future research
directions.

\section{Related work}

\textcolor{rev}{This section reviews previous research from two perspectives: (1) the definition of success, and (2) various types of features related to movie success.} 

\subsection{The definition of success}

We first checked the way in which
success is defined, because it is of paramount importance to the problem.
As we mentioned earlier, past works have focused primarily on gross
box office revenue \citep{Apala2013,Parimi2013,Asur2010,Taylor2012,Mestyan2013,Gopinath2013},
while some used the number of admissions \citep{Baimbridge1997,Meiseberg2013}.
The basic assumption for using the two as success metrics is simple--a
movie that sells well at the box office is considered a success. However,
the two metrics ignore how much it costs to produce a movie. In fact,
our analysis on historical data also found that revenues are not directly
related to profits (more details in Section 5). Thus a more meaningful
measure of success should be profitability, whether it is the numeric
value of profits \citep{Vany1999} or the Return on Investment (ROI)
\citep{Eliashberg2007}. 

After a success metrics was chosen, many studies categorized movies into two
classes based on their revenues (success or not) and adopted binary
classifications for predictions; some considered the prediction as
a multi-class classification problem and tried to classify movies
into several discrete categories, ranging from 'blockbuster' to 'flop'
\citep{Parimi2013}. Meanwhile, there are also predictions on continuous
numerical values of success metrics \citep{Eliashberg2000,Walls2005,Mestyan2013},
with values of these metrics being logarithmized in several studies
\citep{Taylor2012,Vany1999,Zhang2009}. 

\subsection{Features for movie success}
The accuracy of a predictive model depends a lot on the extraction
and engineering of features (a.k.a., independent variables). When
it comes to studying movie success, three types of features have
been explored: audience-based, release-based, and movie-based features.

Audience-based features are about potential audiences' reception of
a movie. The more optimistic, positive, or excited the audiences are
about a movie, the more likely it is to have a higher revenue. Similarly,
a movie with more pessimistic and negative receptions from the public
may attract fewer people to fill seats. Such receptions can be retrieved
from different types of media, such as Twitter \citep{Asur2010},
trailer comments from Youtube \citep{Apala2013}, blogs \citep{Gopinath2013},
new articles \citep{Zhang2009}, and movie reviews \citep{Meiseberg2013}.
The volume of discussions \citep{Asur2010}, the sentiment of review
or comments \citep{Apala2013,Boccardelli2008,Meiseberg2013}, as well
as the star rating from reviews \citep{Bozdogan2013,Taylor2012} have
been used as a means for assessing audience's excitement towards a
movie.

Release-based features focus on the availability of a movie and the
time of its release. One such feature that captures availability at
release is the number of theaters a movie opens in \citep{Zhang2009,Sharda2006,Taylor2012,Parimi2013,Mestyan2013,Walls2005}.
The more theaters that will show a movie, the more likely the movie
will have a higher revenue. Many movies are targeted for releases
at a certain time, or at a time in which they would be eligible to
receive an upcoming award. Holiday release is a feature commonly utilized
in the prediction problem \citep{Gopinath2013,Taylor2012}, as are
seasons and dates of releases (Spring, Summer, etc.) \citep{Gopinath2013,Parimi2013,Bozdogan2013}.
Some studies attempted to capture the competition at the time of release
\citep{Parimi2013,Gopinath2013}, which has the potential of negatively
effecting revenues.

Movie-based features are those that are directly related to a movie
itself, including who are on the cast and what the movie is about. As for cast members, the most popular feature is a movie's star power--whether the movie casts star actors. Star powers of actors have been captured by actor earnings \citep{Parimi2013}, past award nominations \citep{Boccardelli2008}, actor rankings \citep{Taylor2012}, and the number of actors' Twitter followers \citep{Apala2013}. It was agreed that higher star powers are helpful for a movie's success. \textcolor{rev}{However, no research has explored the profitability of actors. As it costs a great amount of money to cast a famous actor, we believe an actor's record of profitability will be a better indicator of a movie's profitability than her record in generating revenues. Moreover, the role of directors in a movie's financial success is often overlooked or downplayed. While some research has investigated the individual success of directors \cite{Lutter2014}, few studies have actually tried to connect directors' star powers to movies' financial success. Some argued that the economic performance of movies is not affected by the presence of star directors \cite{Boccardelli2008}, and directors' values are not as important as actors' values for movie revenues \cite{Meiseberg2008}. As directors play important roles in movie productions \cite{Lutter2014}, this research examined the effect of directors on movie profitability.}

\textcolor{rev}{In addition to individual actors and directors, the cast of a movie has also been explored from a teamwork perspective -- whether individuals in a team can work together and develop ``team chemistry'' \citep{Meiseberg2013}. Studies of organizations and teams have revealed that team members' prior experience or expertise is beneficial for team success, while the diversity of a team helps too, especially in the context of bringing creative ideas and unique experience to teams for scientific research and performing arts \cite{Guimera2005, Uzzi2005}. Both diversity and familiarity of a cast contribute to a director's success in receiving awards \cite{Lutter2014}. Similarly, for movies' financial success, the diversity of a cast is positively correlated with the movie's box-office revenue \citep{Meiseberg2013}, and cast members' previous experience also positively influences revenues \cite{Meiseberg2008}. Nevertheless, there are several important limitations to consider. On one hand, many of the measurements for teamwork were simplistic and problematic. For example, an actor's experience was based solely on the number of previous movie appearances, without considering what types of movies she has contributed to, and thus has more experience in. Also, team members' degree dispersions were used to reflect a team's diversity even though a team composed of actors who have never collaborated with each other can still feature a uniform degree distribution. Although the existence of structural holes can reflect a team's diversity, the measurement of structural holes was simplified to the density of a network. Nevertheless, the two concepts are only very loosely related. On the other hand, the data size was small in many studies. For instance, the top 10 movies (by revenue) in each year (a total sample size of 160-180 movies) were studied in \citep{Meiseberg2008, Meiseberg2013}. With such a small sample, an actor's experience and previous collaboration cannot be completely captured. The selection bias towards more successful movies also hurt the validity of the results. Thus in this research, we leveraged much larger datasets, derived new and more accurate ways to capture individual actors' experience and teams' diversity, and related them to movie profitability.}

In terms of what a movie is about, features such as genre, MPAA rating,
whether or not a movie is a sequel, and run time have often been incorporated
into success predictions. Besides such meta data about a movie,
to get a better idea of a movie's content, one needs to examine its
plot or script. \textcolor{rev}{Two earlier studies leveraged the texts of movie scripts for success predictions \citep{Eliashberg2007,Eliashberg2014}. Some of the basic text-based features are easy to obtain, such as the number of words, number of characters, number of sentences, and the amount of dialogue present. Nevertheless, more informative textual features in these studies depend on manual annotations by human experts, such as the degree to which the story or hero is logical, the degree to which the premise is clear, the degree to which violence is present, and whether or not the story has a believable ending. As movie scripts can be very long, the manual annotations are time-consuming. Also only a small number of movies' scripts are available in a uniform and professional format. Thus a predictive model based on features from scripts can only be trained on a small pool of movies, which may limit its predictive power for future movies. Therefore an automated way to analyze openly available texts about a movie's content is necessary for a decision support system that needs to learn from large-scale datasets.}

For our research question of predicting movie profitability at an
early stage, we cannot take advantages of most audience-based features
and some of the release-based features, as they would not be available when
making investment decisions. For instance, YouTube comments only appear after a movie trailer is released; likewise,
the number of theaters a movie is going to be released in will not
be known until the end of the movies production. \textcolor{rev}{In addition, these features from different groups were treated as standalone and independent, whereas the interaction or match between features from different groups, such as actors' star powers along with their experience with different movie genres, or the popularity of a certain type of movie during a specific time period, can provide valuable information about a movie's success.}

Therefore, we will focus mainly on four types of features: ``Who'' features -- who is involved in a movie,
``When'' features -- when a movie will be released, ``What'' features
from both meta data and texts of movie plot synopses \textcolor{rev}{(movie plot synopses are openly available from most movie data archives, yet they can still reflect movies' content)}, as well as ```Hybrid'' features--the match between ``What'' and ``Who'' and the match between ``What''
and ``When''. \textcolor{rev}{Our feature set includes popular features from the literature (e.g., measuring actor star powers using their total gross revenues), new features proposed to better measure previously proven factors for movie success (e.g., team expertise and diversity), as well as features representing new factors that may be related to movie success (e.g., actor-director collaboration, and market trend by genre).} All the features adopted by our system can be extracted in an automated fashion by using text mining and social network analysis techniques. \textcolor{rev}{In addition, from a theoretical perspective, this study also examined whether previous findings about star powers of actors and directors, and teamwork are still valid when movie success is measured by profit, instead of revenues, based on a much larger dataset}.

\section{The system framework}

\begin{figure}
\includegraphics[scale=0.35]{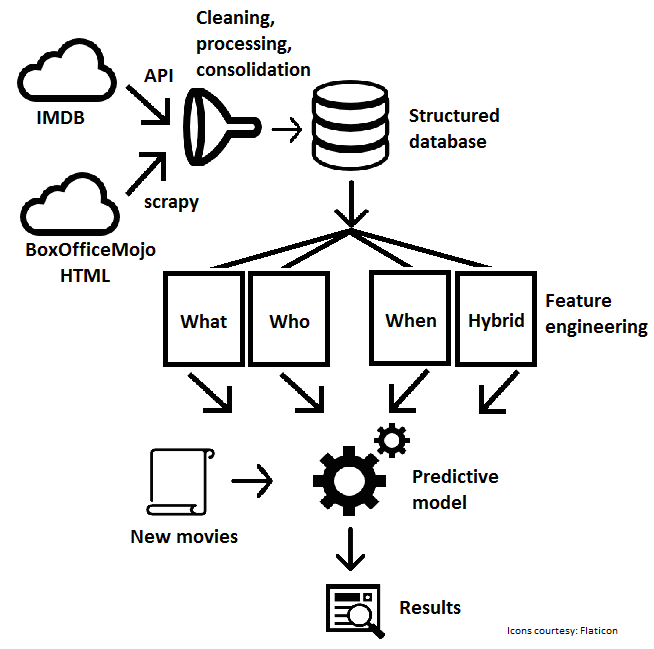} \centering{}\protect\protect\protect\protect\protect\caption{The framework of MIAS.\label{fig:dss}}
\end{figure}

Figure \ref{fig:dss} illustrates the framework of our MIAS. The first
phase is data acquisition, because we based our prediction on historical
data. While there are various online archives of movie data, we picked
two popular and complementary sources -- IMDb and BoxOfficeMojo. IMDb
has better coverage of movie plot synopses, while BoxOfficeMojo, as
its name suggests, provides more comprehensive data of movie revenues
and budgets. In other words, the two data sources can be used jointly
for many movies. As for data collection methods, the two sources are
different as well. IMDb has a convenient API, from which movie data
can be gathered in JSON or XML. The data from BoxOfficeMojo can
only be obtained by the public from its web pages. To get a more comprehensive dataset,
our system employs two scripts: one interacts with APIs and one web
scraper to retrieve and parse HTML data from web pages. We believe
these two methods should be able to handle data from most open archives
on the Internet.

In the second phase, data from both sources is cleaned, transformed,
consolidated, and stored in a database. During this non-trivial undertaking, we need to make sure that acquired features and
texts be put in a consistent format, and the data is not duplicated
within the database. \textcolor{rev}{For example, for movie titles, characters such as ``{*}'' and ``{-}'' are searched for and removed. Such standardization ensures that extraneous characters do not occlude the matching of titles between the two data sources. In the case of plot synopses, the Porter stemmer was used, and stop words (such as ``{the}'') were also removed.}

Phase 3, ``feature engineering'', involves utilizing the acquired
data to construct features that will ultimately be used to train a
predictive model. Categorically, we classify various features into
one of four groups -- ``what'', ``who'', ``when'', or ``hybrid''. Features
used in this study, and the reasons for including them, will be discussed
at length in the next section.

With a reasonable and well-rounded set of features in place, a predictive
model can be trained in Phase 4 of MIAS. Users of MIAS can employ
cross-validation to select optimal parameters (depending upon the
model being tested), as well as for selecting the best performing
method of prediction, and evaluating general system performance. Section
5 will discuss our experiments in detail.

To utilize the predictive model, an investor can take information
provided about a potential investment opportunity and feed it into
MIAS, which would make a prediction based on historical data as to
whether or not it would be profitable.

\section{Feature engineering}

Based on historical data acquired from online archives, we derived four groups of features: ``who'' features, ``what'' features, ``when'' features, as well as ``hybrid'' features that match ``who'' with ``what'', as well as ``what'' with ``when''.

\subsection{``Who'' Features}

\subsubsection{Star Powers}

The very nature of the movie industry is characterized by people who
make movies. Successful actors and directors, such as Brad Pitt, George
Clooney, and Christopher Nolan, are crowd favorites who are well known
throughout the world. Talented individuals such as these can leverage
not only their refined industry skills to make high-quality movies,
but also the associated 'name brand effect', which draws crowds and
increases sales \citep{Asur2010,Elberse2007,Wallace1990}. This effect
is typically referred to as `star power'. \textcolor{rev}{Because our goal is to predict profitability, our star power features for a movie are based on its cast members' records in generating both box-office revenues and profits.}

\begin{enumerate}
\item \textbf{Tenure} of an actor reflects how much experience she/he may
have in the industry. It is calculated as the time difference (in
years) between the movie in which an actor most recently appeared
and that in which he/she first ever appeared. For each movie, we calculate
the \textbf{average }and\textbf{ total} tenure for its first-billed
cast actors. 
\item \textbf{Actor Gross} is about how much revenue an actor has generated
during her/his tenure. Each individual's total gross is the sum of
revenues from all the movies that she has starred in, while an individual's
average gross is her total gross divided by the number of movies she
starred in. For each movie, we calculated the sum and average of total
gross, as well as the average of actor average gross, for all first-billed
cast members. 
\item \textbf{Director Gross }measures the past success of directors. We
calculated for each director the \textbf{total} and \textbf{average}
gross for movies she/he has directed.
\item \textcolor{rev}{\textbf{Actor Profit }measures the amount of profit an actor has earned through his/her career before the movie to be predicted. For each actor, we derived total profit, average profit, and top profit -- the profit of the most profitable movie the actor has appeared in.}
\item \textcolor{rev}{\textbf{Director Profit }represents the amount of profit that a director has earned before the movie to be predicted. Similar to actors, we considered total profit, average profit, and top profit for each director}.

\end{enumerate}

\subsubsection{Network-based Features}

Star power features we listed above reflect whether a movie's cast consists of
senior and successful individuals (actors and the director).
To capture team characteristics, we explored the avenue
of social networks, which have the potential to yield a wealth of
information about inter-personal interactions and collaboration \citep{Zhao2013, Zhao2011}, including teams for movie productions \cite{Meiseberg2013, Lutter2014}.

\begin{figure}
\begin{minipage}[t]{0.35\columnwidth}%
\includegraphics[scale=0.35]{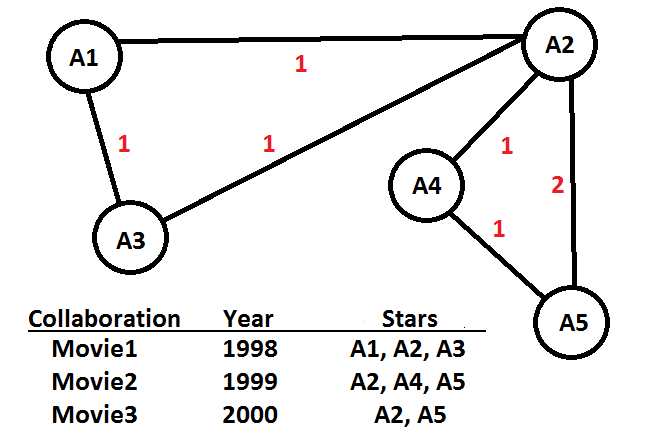}%
\end{minipage}\hfill{}%
\begin{minipage}[t]{0.35\columnwidth}%
\includegraphics[scale=0.35]{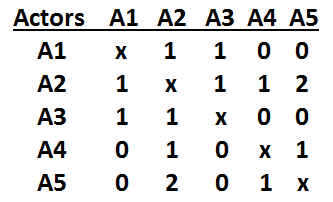}%
\end{minipage}\protect\protect\protect\protect\protect\caption{A collaboration network example with the network structure (left) and the corresponding adjacency matrix (right). \label{fig:collab_net}}
\end{figure}

For our predictive model, we constructed a dynamic collaboration network
among actors based on their co-appearances (i.e. co-starring) in previous
movies. In such a network, a node represents an actor. For
any arbitrary year, an undirected edge was drawn between two actors
if they co-starred in a movie during that year. If an edge already
existed between the two, indicating that they had collaborated in
the past, the edge weight was incremented by 1. Therefore, the
aggregated networks for a given year includes all of the earlier years of collaborations, plus those that happen in that year. Fig \ref{fig:collab_net} shows an example network.

Our network features consist of static features and
dynamic features. When analyzing the team $T_{m}$
for movie $m$ in year $y$, the social network among the movie's
cast members up to year $y-1$ was used to extract the following
static features:

\begin{enumerate}

\item \textbf{Network Heterogeneity: }for each movie, we measured its team diversity by examining the network structural
similarity between cast members. Specifically, based on each actor's
neighborhood vector in the adjacency matrix, we calculated average
cosine similarity between each pair of actors for the movie, which
is denoted by Equation \ref{eq:cos_sim}. In this equation let $|T_m|$
denote the number of cast members in team $T_m$ for movie $m$, $Act_{i}\bullet Act_{j}$
is the dot product between two actors in the team, and $||Act_{i}||||Act_{j}||$
is the magnitude of the two actor vectors. \textcolor{rev}{Higher similarity means team members have been working with similar peers (including one another). Lower similarity suggests higher diversity among team members as their previous collaborators did not have much overlap. We believe this measure can better capture previous collaborations among team members than degree dispersions \cite{Meiseberg2008}, which does not consider who an actor is connected to in a network.}

\begin{equation}
H_m=\frac{1}{(|T_m|(|T_m|-1)/2)}\sum_{i=1}^{|T_m|-1}\sum_{j=i+1}^{|T_m|}\frac{Act_{i}\bullet Act_{j}}{||Act_{i}||||Act_{j}||}\label{eq:cos_sim}
\end{equation}

\item \textbf{Average Degree }represents the average number of unique collaborations
for each cast members in a given movie. This metric is meant to capture
the `degree' to which the team is truly bringing rich expertise
and experience to the production of a movie \citep{Meiseberg2008}. 

\item \textbf{Total and Average Betweenness Centrality: }In addition to
those with many unique collaborators, ``brokers'' who can bridge
between different and otherwise less inter-connected groups are also
at a good position to bring in unique expertise and experience. \textcolor{rev}{These
``brokers'' often have high betweenness centralities and are said
to have high social capital \citep{Burt1992}, and having such ``brokers'' in a team can increase the team's diversity by creating new ideas and producing innovations \cite{Burt2004}.}

\end{enumerate}

\textcolor{rev}{In addition to collaboration among actors, we also considered the collaboration between actors and directors by examining whether an actor and a director have worked together before and whether previous collaborations between them were successful.}

\begin{enumerate}
\item \textcolor{rev}{\textbf{Average Actor-Director Collaboration Frequency} of movie $m$ is average number of times that cast members of $m$ have previously appeared in movies directed by the director of $m$.}

\item \textcolor{rev}{\textbf{Average Actor-Director Collaboration Profitability} of movie $m$ is the average profit per movie earned from all past collaborations between actors and the director of movie $m$.}
\end{enumerate}

Features introduced above are based on the static
structure of the year $y-1$ network before movie $m$ was produced
in year $y$. Once movie $m$ was produced in year $y$, its cast
members formed a new team, which would add edges to the collaboration
network and change its structure. \textcolor{rev}{Our newly developed dynamic network
features tried to capture the spanning of structural holes after a new
movie is produced.}

\textcolor{rev}{Structural holes are an important concept in network analysis \citep{Burt1992}, and networks pertaining to
movies are no exception \citep{Zaheer2007}. Some previous studies believe that a movie, which establishes inter-actor links that span structural holes, is more likely to succeed \citep{Boccardelli2008}, although they did not quantify the degree
to which a movie spans structural holes. Some research used the clustering coefficient for each team member to measure the existence of structural holes \cite{Meiseberg2008}.} The clustering coefficient of a node is the probability that the node's neighbors are also connected to each other. It measures how a node's neighborhood is ``clustered'' together. A network's clustering coefficient is the average of all its nodes' values on this metric. \textcolor{rev}{However, the static value of a team's average clustering coefficient alone in the current collaboration network can only show structural holes at the ego level (i.e., among immediate neighbors of an ego). To capture the spanning of structural holes at the network level, we used the following two dynamic network features to measure how the structure of the collaboration network changed after incorporating the collaboration in a new movie.}

\begin{enumerate}
\item \textbf{Decrease in clustering coefficient: } 

\textcolor{rev}{A new movie will add edges to an existing collaboration network. If these new edges connect nodes that are originally only 2 hops away from each other, then the clustering coefficient of the network will increase, but such edges only reinforce existing clusters. By contrast, if new edges connect nodes that are more than 2 hops away from each other, then clustering coefficients of the two nodes will decrease. For example, node A are node B are 3 hops away from each other. Then a new edge between A and B creates new 2-hop neighbors for both A and B. However, none of these new 2-hop neighbors of A are connected to B, and none of these new 2-hop neighbors of B are connected to A. Otherwise, A and B would have been only 2 hops away from each other to begin with. By creating new neighbors without closing the triad, new edges that connect nodes that are originally far away would decrease the network's clustering coefficient.} It has been found that decreasing the clustering co-efficient of a social network can facilitate the diffusion of information across the network by breaking up existing clusters \citep{Zhao2013}. Thus we also included the decrease in the clustering coefficient of a collaboration network after forming the team for movie $m$ to measure whether new collaborations can break existing clusters.

\item \textbf{Decrease in average shortest path: } We also proposed to use how the production of
a movie $m$ decreases the average shortest path length of the social network, because adding edges that span structural holes usually significantly decreases such path length. Specifically, after adding to $Network_{y-1}$
edges that correspondent to the cast of $T_{m}$ produced at year
$y$, we calculated how much the average shortest path length of the
new network decreased, compared to $Network_{y-1}$. The more such
length decreases, the more movie $m$'s cast can span structural holes
in $Network_{y-1}$. 

\end{enumerate}

\subsection{``What'' Features}

In addition to ``who'' are in the cast, another natural and important
indicator of a movie's future profitability is what the movie is about.
Such information is usually available with high certainty prior to
movie funding efforts. To reflect what a movie is about, the ``what''
features in our model include both meta features, such as \textbf{genre}
(e.g., action, sci-fi, family) and \textbf{rating} (e.g., PG13, and
R), but also fine-grained description of a movie's content--its plot
synopsis.

In text mining, texts from plot synopses can be represented as traditional unigrams
and bigrams, but such representation will have high dimensionality
and, as a result, suffers from sparsity. At a higher level, topic
model techniques, such as Latent Dirichlet Allocation (LDA) \citep{Blei2003},
can give a better picture of what a plot is about. The input for LDA
is a textual corpus from plot synopses and the output is a group of
topics, each being represented by a probabilistic distribution over
words. Those words with high probabilities on a topic are considered
representative keywords for the topic. Each plot synopsis is also
assigned a probabilistic distribution over all the topics. Such \textbf{topic
distribution vector of a movie's plot} reflects the content of the
movie at an aggregated level and can be used as features for predictive
modeling.

In addition to these topics derived from LDA, some movies' plots are
adaptations from other sources, especially when the original sources
had achieved certain levels of success. For example, The Hunger Games
and Harry Potter are both adapted from best-selling novels. As such,
one of our ``what'' features was about \textbf{adaptations}: whether
a movie's plot was adapted from a comic, a true story, or a book/novel.

\subsection{``When'' Features}

With the movie industry being an avenue for entertainment, its market
sees peaks and declines over time, which may speak to how well a pre-production
movie may fare in the future. Thus we incorporated the following ``when''
features in our model: 
\begin{enumerate}
\item \textbf{Average Annual Profit: }is the average profit across all movies
in the year prior to the planned release of movie $m$. This feature
captures the overall profitability of the movie industry before a movie is released. 
\item \textbf{Release dates:} combines several features about when a movie
will be released, including whether it will be a holiday release and
which season of the year (spring, summer, fall, winter). While a holiday
or summer release may attract more of an audience and thus generate
more revenues \citep{Apala2013}, it also requires higher budget for
marketing and distributions during these competitive periods. Although
the exact release date is not completely definitive before filming,
a target trajectory usually exists at the early stage of movie production. 
\end{enumerate}

\subsection{Hybrid Features}

Besides standalone features about ```who'' are in a movie, ``what'' a movie
is about, and ``when'' a movie will be released, it is also
important to capture the ``match'' between these features. Our hybrid
features try to reflect such matches between ``what'' and ``who'',
as well as between ``what'' and ``when''. For example, it may
be important to form a team of actors based on their previous experience
with the genre of the movie being planned, instead of just their star
powers. Similarly, the investment on a movie whose genre is gaining popularity may increase the chance of success.

\subsubsection{``What'' + ``Who''}

In observing the movie industry as a whole, and the actors that tell
the stories, we can distinguish various so-called `roles' that these
actors seem to adopt. For example, Seth Rogan is typified by his appearance
in comedies, and Bruce Willis exhibits a proficiency as an action
movie star. Should a movie then, granted this observation, try to
include those who have extensive experience in its genre? Or conversely, does a surprising
cast draw a greater audience to theaters (e.g., having Bruce Willis
in a comedy or having action star Arnold Schwartzenegger in a romantic
love story)? \textcolor{rev}{Although these questions have not been addressed in the literature, we believe that better measurements of an actor's expertise with regard to movie genres can help us more accurately determine the expertise and diversity of a movie's cast.}

To measure an actor's previous experience and expertise in movies
with different genre we define, for each actor $j$, a
genre experience vector $A_{j}=[a_{j,1},...,a_{j,k},...,a_{j,K}]$,
where $a_{j,k}$ is the proportion of the number of times actor
$j$ appeared in movies with genre $k$. A total of $K=26$ unique
genres are defined.

Similarly, a movie $m$ is also represented as a genre vector $G_{m}=[g_{m,1},...,g_{m,k},...,g_{m,K}]$,
where $g_{m,k}=1$ indicates that movie $m$ has genre $k$, and $g_{m,k}=0$
otherwise. Note that some movies can have more than one genre. For
example, the genre of `Spiderman' is both action and adventure.

By measuring the similarity between actors' genre experience vectors
and movies' genre vectors, we designed several features that speak to
the genre-based expertise brought by cast members to a given film
$m$'s team $T_{m}$. 
\begin{enumerate}
\item \textbf{Average Genre Expertise (AGE): }captures the average cast
experience with respect to the current movie's genre. Movie $m$'s
AGE is defined in Equation \ref{eq:avg_gen_exp}.

\begin{equation}
AGE_{m}=\frac{1}{|T_m|}\sum_{j=1}^{|T_m|}{G_{m}}\bullet{A_{j}}\label{eq:avg_gen_exp}
\end{equation}

\item \textbf{Weighted Average Genre Expertise (WAGE) }is an extension of
AGE. AGE only considers the frequency that each actor starred in a
genre, while WAGE goes a step further to incorporate an actor's star
power, measured by actor gross, in each genre. As defined in Equation \ref{eq:weight_avg_gen_exp},
WAGE of movie $m$ is essentially the movie's AGE weighted by each
cast member $j$'s gross revenue $R_{j}$. In other words, a movie
with a big star who is familiar with its genre will have high WAGE.

\begin{equation}
WAGE_{m}=\frac{1}{|T_m|}\sum_{j=1}^{|T_m|}log(R_{j})*({G_{m}}\bullet{A_{j}})\label{eq:weight_avg_gen_exp}
\end{equation}

\item \textbf{Cast Novelty} is \textcolor{rev}{defined in a way similar to WAGE. While WAGE is an average value that tries to capture a cast's experience in the movie's genre, cast novelty focuses on team diversity--whether a
movie has a big star who has rarely appeared in movies of this genre
before.} It is the maximum value among all actors' star-power-weighted
inverse experience in movie $m$'s genre (Equation \ref{eq:cast_novelty}). \textcolor{rev}{The higher the value is, the more diverse the cast is, in terms of having an unexpected star appearing in a given movie.}

\begin{equation}
CN_{m}=max\{\frac{log(R_{j})}{{G_{m}}\bullet{A_{j}}+1},\forall j\in T_{m}\}\label{eq:cast_novelty}
\end{equation}

\end{enumerate}

\subsubsection{``What'' + ``When''}

Similar to the overall market volume for movies, which may change from year to
year, consumers' preferences of movies may also evolve with time.
For example, while movies like ``American Pie'' and ``National
Lampoon's Van Wilder'' were popular in the late 90's and early 2000's,
movie-goers recently have been flocking to horror movies, such as
``Paranormal Activity'', and those characterized by superheroes,
such as ``The Avengers'' and ``Captain America''. Although the
latter category is nothing definitively new to the silver screen,
the movie industry has seen greater levels of success in recent years
with this particular focus and, as such, a greater influx of such
movies. \textcolor{rev}{Meanwhile, competitions may also affect the profitability of movie $m$ because other movies released during a similar time period may detract from movie $m$'s viewer-base \citep{Parimi2013}. Thus, in addition to capturing ``when'' a movie will be released, we also consider how movies with similar genre performed in the previous year, as well as the level of competition during a movie's planned release time.}

\begin{enumerate}
\item \textbf{Annual Profitability Percentage by Genre }is the percentage
of profitable movies, which have the same genre with movie $m$, in
the year prior to the planned release of movie $m$. This feature
reflects the degree of success for movies that share the same genre
as the movie being considered. 

\item \textbf{Annual Weighted Profitability by Genre (AWPG)} is derived
from movie genre vectors defined earlier in this paper. For movie
$m$ in year $y$, the profitability of each movie ${m}^{'}$ in year
$y-1$ are summed up and weighted by the cosine similarities between
genre vectors of $m$ and ${m}^{'}$. Equation \ref{eq:ann_weigh_prof_gen}
illustrates how to calculate the AWPG for movie $m$ in year $y$,
where $G_{m}$ is the genre vector for movie $m$ and $p({m}^{'})$
is the profitability of movie ${m}^{'}$. This feature indicates the
overall previous-year profitability of movies whose genre is similar
to a given movie. 
\begin{equation}
AWPG_{m}=\sum_{{m}^{'}\epsilon y-1}sim(G_{m},G_{{m}^{'}})*p({m}^{'})\label{eq:ann_weigh_prof_gen}
\end{equation}
\item \textcolor{rev}{\textbf{Competition} reflects what other movies will be released during a similar time period. It is calculated by considering the average star-power of all other movies released within 1 month of movie $m$'s release date. This feature indicates the degree to which other big-name stars appearing in movies that may detract from movie $m$'s viewership.} 

\end{enumerate}

\section{Experiments}

\subsection{Dataset and Basic Statistics}

Our original dataset, collected from both BoxOfficeMojo and IMDb,
consisted of 14,097 movies, along with 4,420 actors. While movies
in our dataset date back to 1921, we focused our study to movies released
during the 11-year period of 2000-2010 (inclusive), because this period
is recent enough to reflect the current state of the industry and
there has been a sufficient amount of elapsed time since movies' release
for revenue data to be accurately updated.

As our goal is to predict movie success measured by profits, our dataset
for experiments only included those movies that have both budget and
box office revenue data available. We also excluded movies with an
`Unknown' genre, or an `Unknown' MPAA rating. Movies with `Documentary'
genre were also excluded, as those are typically not released to theaters
and may not involve professional actors. Additionally, any
\textcolor{rev}{movie designated as being part of a franchise, a sequel, or a remake was also excluded (e.g. Iron Man, Iron Man 2, etc.). We made this decision because the success of a sequel can depend heavily on the success of earlier movies in the same franchise. Also, the content of sequels and remakes and their selections of cast members are also highly limited by earlier counterparts. Thus what is behind the success of a sequel or remake may be very different from that of other movies.}

\begin{figure}
\centering{}\includegraphics[scale=0.75]{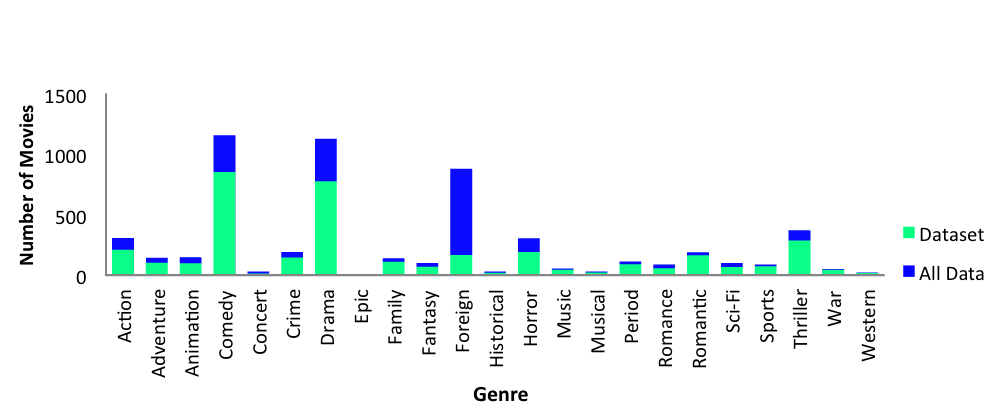}\protect\protect\protect\protect\protect\caption{Distribution of movies by genre (2000-2010).\label{fig:distrib_movies_genre}}
\end{figure}

With all of these considerations in mind, our final dataset for experiments
consisted of 2,506 movies. A distribution by genre of these 2,506
movies, relative to all movies released during the period, is related
in Figure \ref{fig:distrib_movies_genre}. The distribution suggests
that our dataset is a representative sample overall, with the exception
of the `Foreign' genre. This makes sense because budget and revenue
data may be more difficult to obtain for movies that are produced,
and in all likelihood, released outside the U.S. Based on the plot
synopses of these movies, we used LDA to generate 30 topics. Top keywords
of these topics are listed in Table \ref{tab:LDAtopics}.

\begin{table}[htbp]
\centering %
\begin{tabular}{|c|l|}
\hline 
\textbf{Topic}  & \textbf{Keywords} \tabularnewline
\hline 
1  & Wife, husband, marriage, child, couple \tabularnewline
\hline 
2  & People, movie, story, show, tv \tabularnewline
\hline 
3  & Back, even, good, time, start \tabularnewline
\hline 
4  & Music, band, famous, star, place \tabularnewline
\hline 
5  & First, world, people, state, country \tabularnewline
\hline 
6  & Money, back, plan, help, deal \tabularnewline
\hline 
7  & Man, begin, believes, situation, hospital \tabularnewline
\hline 
8  & Life, young, city, world, lives \tabularnewline
\hline 
9  & Find, way, help, search, journey \tabularnewline
\hline 
10  & Group, find, survive, crew, remote \tabularnewline
\hline 
11  & One, life, never, day, always \tabularnewline
\hline 
12  & World, stop, evil, power, battle \tabularnewline
\hline 
13  & Family, father, son, mother, home \tabularnewline
\hline 
14  & Night, day, car, trip, train \tabularnewline
\hline 
15  & New, life, dream, everything, lost \tabularnewline
\hline 
16  & Man, young, become, past, truth \tabularnewline
\hline 
17  & He, want, she, know, tell \tabularnewline
\hline 
18  & War, mission, American, government, fight \tabularnewline
\hline 
19  & Love, young, woman, heart, marry \tabularnewline
\hline 
20  & Team, game, win, dream, big \tabularnewline
\hline 
21  & Work, job, business, company, success \tabularnewline
\hline 
22  & School, high, parents, boy, girl \tabularnewline
\hline 
23  & Friend, girlfriend, party, boyfriend, college \tabularnewline
\hline 
24  & Story, film, based, documentary, history \tabularnewline
\hline 
25  & Police, murder, drug, prison, kill \tabularnewline
\hline 
26  & Two, lives, relationship, together, sex \tabularnewline
\hline 
27  & He, find, she, finally, arrive \tabularnewline
\hline 
28  & Years, time, later, death, since \tabularnewline
\hline 
29  & Events, forced, act, unexpected, secrets \tabularnewline
\hline 
30  & Town, local, small, gang, store \tabularnewline
\hline 
\end{tabular}\protect\caption{Topics and keywords generated by LDA from plot synopses.}
\label{tab:LDAtopics} 
\end{table}

While the experiment will predict the success of 2,506 movies during
a 11-year period, the collaboration network we built for this study
incorporates the collaboration between all the actors in all the 14,097
movies in our dataset. The initial unweighted, undirected network
was aggregated to the year 1999, with networks for subsequent years
being updated to reflect that year's new collaborations. In all, we
created 11 snapshots of the collaboration network from 1999 to 2009.

\subsection{The measure of success}

\begin{figure}
\centering{}\includegraphics[scale=0.5]{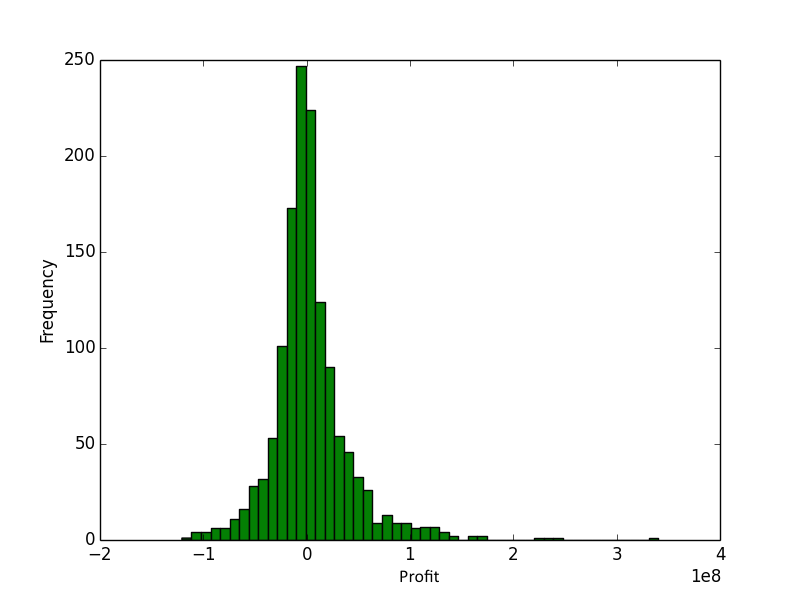}\protect\protect\protect\protect\protect\caption{Distribution of raw profits (in USD) for movies in our experiment
dataset.\label{fig:profit_dist}}
\end{figure}

To predict whether a movie is successful, we first need to measure
a movie's profitability. With both revenue and budget data for each
movie in our dataset, we can certainly calculate the raw profit value
(i.e., revenue-budget) for each movie. The distribution of raw profit
values in Figure \ref{fig:profit_dist} appears to follow a normal
distribution--a large number of movies have near-zero profits. With
only 36\% of the movies in our dataset having positive profits, investors
do need a decision support system to help them pick the right movie
to invest in.

Nevertheless, while using profit values is intuitive, gaining a profit
of \$10,000 from a movie that costs \$1 million to produce is certainly
not an attractive investment. Thus in our experiments, we adopted
a popular metric of profitability--\textit{return on investment (ROI)}
as in \citep{Eliashberg2007}. It considers both profit and budget
and is defined in Equation \ref{eq:profMargin}. The higher the ROI
is, the more profitable a movie is, and vice versa.

\begin{equation}
ROI=\frac{Revenue-Budget}{Budget}\label{eq:profMargin}
\end{equation}

Interestingly, the data suggests that profitability, as measured by ROI, is not necessarily reflected by box office revenues. The correlation coefficient between revenues and ROIs is only \textit{0.077}. In other words, having a great box office revenue does not necessarily mean a high ROI. This further highlights the need for the accurate prediction of profitability.



\subsection{Classification of profitability}

The prediction of a movie's success can be considered as a classification
problem, where movies are classified into discrete bins or classes. Then
the problem is to decide whether a movies should be considered a success
or not based on ROI. Although there is no agreed-upon industry `gold
standard' as to an ideal ROI (other than that `higher is better'),
it is reasonable to assume that one would like to see some substantive
returns from a successful movie, given that millions of dollars are
invested with considerable risks. Also, any form of profit is better
than a loss. With these in mind, we elected to define the decision
boundary between successful and unsuccessful movies in two different ways
for both binary and multi-class predictions.

\textcolor{rev}{For both binary and multi-class classification of movie success, we tried a variety of algorithms, including logistic regression, naive Bayesian, support vector machines (SVM), multilayer perceptron (MLP), decision trees (J48), random forest, and the LogitBoost algorithm, and selected the algorithm with the best overall performance based on the following six metrics (all results were obtained using 10-fold cross-validation), where higher values indicate better performance.} 

\begin{enumerate}
\item The Area under the Receiver Operator Characteristic curve (AUC). The
Receiver Operator Characteristic curve plots the true positive rate
(the Y-axis) against the false positive rate (the X-axis). An AUC
of 1 means a perfect classification while 0.5 refers to a random guess.
Being more robust against prior distributions, AUC is considered by
many to be one of the best indicators of a classifier's performance. \textcolor{rev}{In multi-class classifications, we reported a weighted average of AUCs. This is computed by calculating the AUC obtained for each class, which is derived by making predictions as to whether an instance is in the currently considered class or not the currently considered class (1 vs.~all), and then weighting each AUC according to the number of instances that fall into each of these classes relative to the total number of instances.}

\item Classification accuracy, which is the percentage of correctly
predicted instances. 

\item \textcolor{rev}{Precision (positive class), which is the number of instances classified as being positive that are actually successful, divided by the number of instances classified as being successful.}

\item \textcolor{rev}{Recall (positive class), which is the number of instances classified as being positive that are actually successful, divided by the number of instances that are actually successful.}

%
\end{enumerate}

\textcolor{rev}{In addition to identifying the best performing algorithm, we also
evaluated whether, and how, features we proposed in this research contributed to the prediction. We included into a `New' feature group those novel features which we proposed and are used for the first time to predict movie success. Features in this `New' group include (1) features related to actor and director profits, actor-director collaboration, dynamic network features (e.g., decrease in the average shortest path length) from the `Who' group; (2) topic distribution features from the `What' group; (3) average annual profit in the `When' group; and (4) all features in the `Hybrid' group.}


\textcolor{rev}{To further evaluate the performance of our predictive model, we also compared it with two benchmark models -- Benchmark 1 was based on \citep{Vany1999} and Benchmark 2 was based on \citep{Walls2005}. Among previous studies of box office revenue predictions, we selected these two because most features used in their studies are available prior to a movie's production, which is similar to our early prediction problem. For example, the following features were used in \citep{Vany1999}: star power, sequel, genre, rating, and year of release. Similarly, \citep{Walls2005} used film budget/cost, number of screens the film was released on, sequel, star power, genre, and rating. We excluded the sequel feature, as our dataset excludes such movies, and the number of screens feature, because this information is only available before the release. To make the comparison consistent, we used our definition of star power for the two benchmark models. We reported results of best-performing classifiers for both benchmark methods, along with the performance of our approach.}

\subsubsection{Binary classification}

In the case of binary classifications, a movie is classified into
one of the two classes: successful or unsuccessful movies. Two decision
boundaries are evaluated and both ensure a sufficient amount of ROI
is garnished if a movie is considered successful.

\begin{enumerate}
\item The first decision boundary is that a movie is considered successful
if its ROI is within top 30\% of all movies. For our dataset, this
threshold of profitability translates to $ROI$$\geq24\%$. The performance of the top two classification algorithms is listed in Table \ref{tab:profit_top_30}, \textcolor{rev}{with a Random Forest classifier (n=200) generating the highest AUC, accuracy, and recall, and a LogitBoost classifier leading in precision.}

\begin{table}[htbp]
\centering %
\begin{tabular}{|l|c|c|c|c|}
\hline 
Classifier  & \multicolumn{2}{c|}{Random Forest} & \multicolumn{2}{c|}{LogitBoost} \tabularnewline
\hline 
Model  & Full model & w/o New Features & Full model & w/o New Features \tabularnewline
\hline 
AUC  & 0.863 & 0.616 &  0.833 & 0.653\tabularnewline
\hline 
Accuracy  & 0.834  &0.675 & 0.812 & 0.697  \tabularnewline
\hline 
Precision & 0.82 &0.454  & 0.844  & 0.492 \tabularnewline
\hline 
Recall & 0.575  & 0.380 & 0.465 & 0.129 \tabularnewline
\hline
\end{tabular}\protect\caption{Top 2 prediction results of our binary classification model and the performance without 'New'
features (with top 30\% ROIs as the decision boundary.}
\label{tab:profit_top_30} 
\end{table}

\textcolor{rev}{Table \ref{tab:profit_top_30} also lists the performance of the top 2 classifiers when `New' features were removed. As can be observed, AUC and accuracy of the classifier deteriorate 29\% and 20\% respectively for random forest, and 22\% and 15\% for LogitBoost. Precision and recall also drops greatly after `New' features were removed. All of these highlighted the contribution of `New' features to the prediction. In addition, top 2 classifiers of the two benchmark models (Table \ref{tab:benchmark_30_roi}) trail our model in all of the four performance metrics. For instance, AUCs of the two benchmark models are respectively 19\% and 25\% lower than that of ours.}

\begin{table}[htbp]
	\centering %
	\begin{tabular}{|l|c|c|c|c|}
		\hline
		& \multicolumn{2}{c|}{Benchmark 1} & \multicolumn{2}{c|}{Benchmark 2} \tabularnewline
		\hline 
		Classifier  & \multicolumn{1}{c|}{Logistic Regression} & \multicolumn{1}{c|}{Naive Bayesian} &\multicolumn{1}{c|}{Logistic Regression} & \multicolumn{1}{c|}{LogitBoost} \tabularnewline
		\hline 
		AUC  & 0.672  & 0.651 & 0.701  & 0.651 \tabularnewline
		\hline 
		Accuracy  & 0.702  & 0.686 & 0.724  & 0.686 \tabularnewline
		\hline 
		Precision & 0.516  & 0.475 & 0.603  & 0.475  \tabularnewline
		\hline 
		Recall  & 0.188  & 0.367 & 0.252  & 0.367  \tabularnewline
		\hline 
	\end{tabular}\protect\caption{Top 2 prediction results for benchmark binary classification models (with top 30\% ROIs as the decision boundary).}
	\label{tab:benchmark_30_roi} 
\end{table}

\item The second boundary we tested is $ROI\geq67\%$, which corresponds
to 1/4 standard deviation above the mean ROI. With this threshold,
21.4\% of the movies in our dataset were considered successful. Compared
to the decision boundary of top 30\% ROI, this boundary further raises
the bar for a movie to be successful. By defining profitability in
this manner, our predictive task has become easier, which was evidenced by an increase in model performance. Top performing algorithms are able to reach AUC and accuracy over 0.9 (see Table \ref{tab:quart_st_dev_profit}). 

\textcolor{rev}{At the same time, our `New' feature still make great contribution to the classification--the removal of these features dropped the AUC of the best-performing random forest classifier by 24\%, and the LogitBoost classifier's AUC decreased by 20\%. Similar decrease can be found for accuracy, precision, and recall. Our model also keeps the advantage over the two benchmark models, leading by 22\%-27\% on AUCs.}

\begin{table}[htbp]
\centering %
\begin{tabular}{|l|c|c|c|c|}
\hline 
Classifier  & \multicolumn{2}{c|}{Random Forest} & \multicolumn{2}{c|}{LogitBoost} \tabularnewline
\hline 
 Model & Full model & w/o New Features & Full model & w/o New Features \tabularnewline
\hline 
AUC  & 0.921  & 0.707 & 0.917  & 0.735\tabularnewline
\hline 
Accuracy  & 0.904  & 0.749 & 0.891   & 0.796\tabularnewline
\hline 
Precision  & 0.874 &0.399  & 0.855 & 0.583\tabularnewline
\hline 
Recall  & 0.646  & 0.338 & 0.593   & 0.164\tabularnewline
\hline 
\end{tabular}\protect\caption{Top 2 prediction results of our binary classification model and the performance without `New' features (with $ROI\geq67\%$ as the decision boundary).}
\label{tab:quart_st_dev_profit} 
\end{table}

\begin{table}[htbp]
	\centering %
	\begin{tabular}{|l|c|c|c|c|}
		\hline
		& \multicolumn{2}{c|}{Benchmark 1} & \multicolumn{2}{c|}{Benchmark 2} \tabularnewline
		\hline 
		Classifier  & \multicolumn{1}{c|}{Logistic Regression} & \multicolumn{1}{c|}{LogitBoost} &\multicolumn{1}{c|}{Logistic Regression} & \multicolumn{1}{c|}{LogitBoost} \tabularnewline
		\hline 
		AUC  & 0.754  & 0.726 & 0.756  & 0.725 \tabularnewline
		\hline 
		Accuracy  & 0.786  & 0.795 & 0.793  & 0.761 \tabularnewline
		\hline 
		Precision & 0.500  & 0.597 & 0.547  & 0.436  \tabularnewline
		\hline 
		Recall  & 0.175  & 0.132 & 0.194  & 0.397  \tabularnewline
		\hline 
	\end{tabular}\protect\caption{Top 2 prediction results for benchmark binary classification  models (with $ROI\geq67\%$ as the decision boundary).}
	\label{tab:benchmark_quart_st_dev_profit} 
\end{table}

%

\end{enumerate}

\subsubsection{Multi-class classification}

In the case of multi-class classifications, we defined three possible
classes for a movie: positive (`success'), negative (`failure'), or
neutral (`average') to provide more information to investors on where
they could expect a movie to fall as far as profitability is concerned.
For the multi-class prediction, we explored the imposition of cost
associated with mis-classification, because the three classes are
ordinal, whereas binary classes are nominal. In other words, not all
mis-classification errors are equally severe. For example, for investment
decision support, predicting a failure to be a success would be worse
than predicting it to be a neutral movie. The cost matrix for the
multi-class classification is in Table \ref{tab:cost_matrix}--the
penalty imposed for classifying a $successful$ movie as $failure$
is $2$, and vice-versa, whereas the penalty for only mis-classifying
by one ordinal category (i.e., success as neutral, neutral as failure,
etc.) is $1$.

\begin{table}[htbp]
\centering %
\begin{tabular}{|l|c|c|c|}
\hline 
Actual/Predicted  & Positive  & Neutral  & Negative \tabularnewline
\hline 
Positive  & 0  & 1  & 2 \tabularnewline
\hline 
Neutral  & 1  & 0  & 1 \tabularnewline
\hline 
Negative  & 2  & 1  & 0 \tabularnewline
\hline 
\end{tabular}\protect\caption{The cost matrix used in multi-class classification}
\label{tab:cost_matrix} 
\end{table}

Similar to binary classifications, we defined the three classes of success in two ways as well. 
\begin{enumerate}
\item The first way was to split movies into three equal-sized classes:
the positive class consists of movies with top 1/3 ROIs ($ROI\geq10\%$),
the negative class consists of movies with the bottom 1/3 ROIs ($ROI\leq-78\%$),
and the other middle 1/3 into the neutral class ($-78\%<ROI<10\%$).

\item The second way was to classify movies with top 1/4 ROIs as positive
($ROI\geq47\%$), the bottom 1/4 ROIs as negative ($ROI\leq-91\%$),
and the rest as neutral $-91\%<ROI<47\%$). In other words, after
dividing movies into 4 equal-sized groups, the top and bottom groups
become their own classes, and the two groups in the middle (i.e.,
half the movies) were merged into the neutral class.
\end{enumerate}

\begin{table}[htbp]
\centering %
\begin{tabular}{|l|c|c|c|c|}
\hline 
Measure  & \multicolumn{2}{c|}{The 1st decision boundary} & \multicolumn{2}{c|}{The 2nd decision boundary} \tabularnewline
\hline
 Model & Full model & w/o New Features & Full model & w/o New Features\tabularnewline
\hline 
AUC  & 0.847  & 0.636 & 0.85 &0.657  \tabularnewline
\hline 
Accuracy  & 0.679  & 0.459 & 0.73 & 0.508  \tabularnewline
\hline 
Precision (Pos. Class)  & 0.769 & 0.483 & 0.803 & 0.435  \tabularnewline
\hline 
Recall (Pos. Class)  & 0.711 & 0.482 & 0.671  & 0.424 \tabularnewline
\hline 
Total cost  & 986 & 1882 & 732 & 1505  \tabularnewline
\hline 
\end{tabular}\protect\caption{Multi-class classification results of our model from the best-performing random forest classifier, and the performance without `New' features.}
\label{tab:multi_class_pred} 
\end{table}

\begin{table}[htbp]
\centering %
\begin{tabular}{|l|c|c|c|c|}
\hline 
Measure  & \multicolumn{2}{c|}{The 1st decision boundary} & \multicolumn{2}{c|}{The 2nd decision boundary} \tabularnewline
\hline
 Model & Benchmark 1 & Benchmark 2 & Benchmark 1 & Benchmark 2\tabularnewline
\hline 
AUC  & 0.77  & 0.626 & 0.806 & 0.657  \tabularnewline
\hline 
Accuracy  & 0.578  & 0.448 & 0.651 & 0.508  \tabularnewline
\hline 
Precision (Pos. Class)  & 0.474 & 0.456 & 0.473 & 0.406  \tabularnewline
\hline 
Recall (Pos. Class)  & 0.452 & 0.467 & 0.362  & 0.383 \tabularnewline
\hline 
Total cost  & 1534 & 1915 & 1140 & 1509  \tabularnewline
\hline 
\end{tabular}\protect\caption{Multi-class classification results of benchmark models using random forest classifiers.}
\label{tab:multi_class_benchmark} 
\end{table}

After comparing the performance of several classification models (\textcolor{rev}{including J48, Naive Bayesian, MLP, SVM, logistic regression, and LogitBoost}), random forest still emerged as the best classifier for both decision boundaries. \textcolor{rev}{Table \ref{tab:multi_class_pred} lists its performance measures and Table \ref{tab:multi_class_benchmark} shows the performance from the two benchmark methods. Similar to that of binary classifications, our model outperforms the two benchmark models by reducing the total mis-classification cost by 36\%-52\%. Meanwhile, `New' features keep making great contributions to the prediction--the lack of these features from our model can double the mis-classification cost.}




\section{Discussions}

\subsection{Important factors for movie profitability}

\textcolor{rev}{While a random forest classifier can do a good job in predicting whether a movie will be successful, it is also
important to understand factors behind such success. A regression model would help us better assess the degree to which individuals features influence predictive results, and to examine whether they are indicative of movie profitability. Besides, a regression model can also provide predictions on numeric values, in case the classification of movies into 2 or 3 discrete groups is not sufficient for investors' needs. Thus we also explored predicting continuous ROI values. It is worth noting that because the distribution of ROI is highly skewed, we applied a logarithm transformation to ROI, in the format of $log(ROI+1)$ as in \citep{Eliashberg2007}. }

\subsubsection{Regression analysis}

\textcolor{rev}{We tried 6 different algorithms, namely LASSO, Support Vector Regression (SVR), Ridge Regression, CART, M5P Trees, and REP Tree. Among them, we were particularly interested in LASSO and Ridge Regression for two reasons: First, coefficients of each feature in these models are able to offer valuable insights into how each feature contributes to a movie's profit. Second, multi-collinearity may exist among our features (or independent variables). For example, the correlation between Total Actor Profit and Average Actor Profit is 0.96 (p-value<0.001). Such collinearity could negatively affect the validity of regular regression models, such as OLS regression. By contrast, LASSO and Ridge Regression both use regularization ($L1$ and $L2$ respectively) to penalize the non-zero values of the regression coefficients. Such regularization allows the model to select features that are more informative for the prediction and reduce the impact of collinearity \cite{Kuhn_2013}.} 

\textcolor{rev}{
\begin{table}[htbp]
	\centering %
	\begin{tabular}{|l|c|c|c|c|c|c|}
		\hline 
		\diaghead{\theadfont Diag CHeadss  II}{Measure}{Algorithm}  & LASSO & SVR & Ridge Regression & CART & M5P Tree & REP Tree\tabularnewline
		\hline 
		RMSE  & 0.878  & 1.180 & 1.10 & 1.232 & 0.906 & 0.929 \tabularnewline
		\hline 
	\end{tabular}\protect\caption{Results for predicting Log(ROI+1) using various algorithms (10-fold cross validation.}
	\label{tab:log_roi_results} 
\end{table}
}

\textcolor{rev}{Table \ref{tab:log_roi_results} compares root mean squared errors (RMSE) of the 6 algorithms, with LASSO being the best one in predicting numeric values of ROI. Thus we used coefficients from LASSO to reveal factors behind movie success. To obtain these coefficients, we iteratively increased the penalizing $\lambda$ value until all attribute-wise variance inflation factors are reduced to below 10 \citep{VIF}. After achieving such a result with $\lambda=0.0065$, 48 out of the 120 features in the LASSO model ended up with non-zero coefficients: 16 have negative coefficients, and 32 have positive coefficients.}

\begin{table}[htbp]
	\centering %
	\begin{tabular}{|l|c|}
		\hline 
		Feature group & Number of features \tabularnewline
		\hline
		Who (Star power) & 7 \tabularnewline
		\hline 
		Who (Network-based) & 4 \tabularnewline
		\hline 
		What & 24 \tabularnewline
		\hline 
		When &  2 \tabularnewline
		\hline 
		Hybrid (What + Who) &  9 \tabularnewline
		\hline 
		Hybrid (What + When) & 2 \tabularnewline
		\hline
		`New' features & 27 \tabularnewline
		\hline
	\end{tabular}\protect\caption{Number of features from each feature group for the 48 features with non-zero coefficients in LASSO.}
	\label{tab:feature_group_dist} 
\end{table}

\textcolor{rev}{Table \ref{tab:feature_group_dist} lists how many of the 48 features are from each feature group, including the `New' feature group for novel features proposed in this research. It turns out the 48 features cover all feature groups, and more than half of them are `New' features. Besides the `New' feature group, the `What' group contributes the most features, mainly because the group has many features that are not necessarily correlated. For example, 12 out of the 30 topics derived from LDA are among the 48. }

\textcolor{rev}{We also listed the top 5 features by the value of their coefficients (positive and negative) in Tables \ref{tab:reg_coeff_rank_pos} and \ref{tab:reg_coeff_rank_neg}. The top 5 features having positive coefficients are dominated by Star power features from the `Who' group. In addition, those released in winter are more likely to earn higher profits, and the success of movies with the same genre in the previous year is also positively correlated with a movie's profit. Meanwhile, those with top negative coefficients are all `What' features, including genre (drama and foreign), `R' rating, and plot topics related to wars and music.}

\begin{table}[htbp]
	\centering
	\begin{tabular}{|l|l|c|}
		\hline
		\textbf{Feature group} & \textbf{Feature}     & \textbf{Coefficient} \\ \hline
		Who (Star power)    & Avg. profit of actor-director collaboration*   &  0.143 \\ \hline
		Who (Star power) & Avg. Director Gross &  0.039  \\ \hline
		When & Winter Release  &  0.036  \\ \hline
		Who (Star power) & Total Actor Profit*   &  0.035 \\ \hline
		Hybrid (What + When) & Annual Profit \% by Genre* & 0.033 \\ \hline
	\end{tabular}
	\caption{Top 5 features with the highest positive regression coefficients from the LASSO model (* designates a `New' feature).}
	\label{tab:reg_coeff_rank_pos} 
\end{table}

\begin{table}[htbp]
	\centering
	\begin{tabular}{|l|l|c|}
		\hline
		\textbf{Feature group} & \textbf{Feature}     & \textbf{Coefficient} \\ \hline
		What   & R rating   &  -0.058 \\ \hline
		What & Drama Genre &  -0.012  \\ \hline
		What & Topic 18 (war, mission, American, government, fight)* & -0.012  \\ \hline
		What & Topic 4 (music, band, famous, star, place)* & -0.011 \\ \hline
		What & Foreign Genre & -0.009 \\ \hline
	\end{tabular}
	\caption{Top 5 features with the lowest negative regression coefficients from the LASSO model (* designates a `New' feature).}
	\label{tab:reg_coeff_rank_neg}
\end{table}


\subsubsection{Star powers and movie profits}

\begin{figure}
\centering{}%
\begin{minipage}[t]{0.45\columnwidth}%
\includegraphics[scale=0.3]{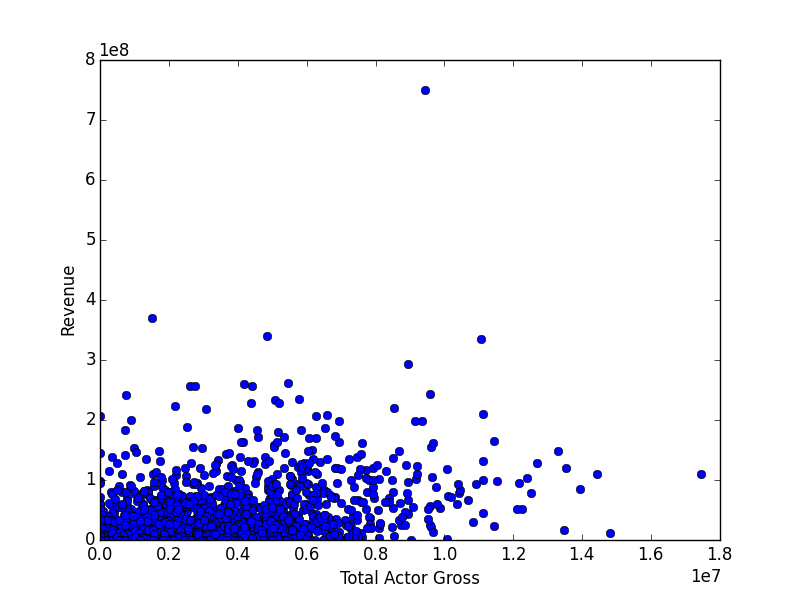}%
\end{minipage}\hfill{}%
\begin{minipage}[t]{0.45\columnwidth}%
\includegraphics[scale=0.3]{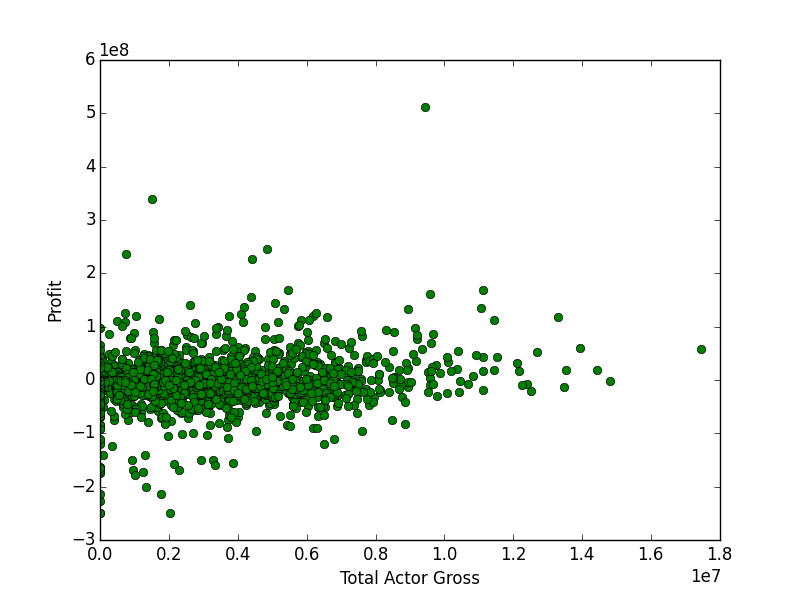}%
\end{minipage}
\protect\protect\protect\protect\protect\caption{Total actor gross vs. movie revenues (left) and profits (right).\label{fig:star_vs}}
\end{figure}

\textcolor{rev}{As we mentioned in the previous paragraph, star power features have a large and positive bearing on the success of movies. While previous studies agreed that higher start powers are generally associated with movie success \citep{Vany1999}, they relied on movies' box-office revenue with actors' star power measured by their total gross. Figure \ref{fig:star_vs} plotted total actor gross against movie revenues and profits in our experiments. As we can observe, although total actor gross is moderately correlated with movie revenues (Pearson correlation coefficient \textit{0.46}), the correlation is much weaker with profits (Pearson correlation coefficient \textit{0.16}). In other words, having actors who have earned big box-office revenues in a movie does not necessarily mean more profits for the movie. Such difference also further highlights the difference between measuring movie success with revenues and profits.}

\begin{figure}
	\centering{}%
	\begin{minipage}[t]{0.45\columnwidth}%
		\includegraphics[scale=0.3]{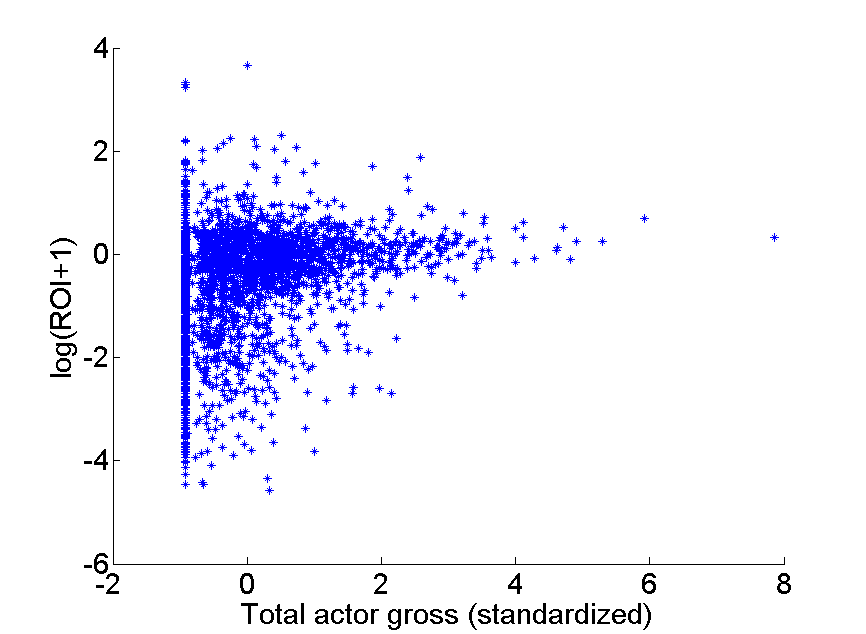}%
	\end{minipage}\hfill{}%
	\begin{minipage}[t]{0.45\columnwidth}%
		\includegraphics[scale=0.3]{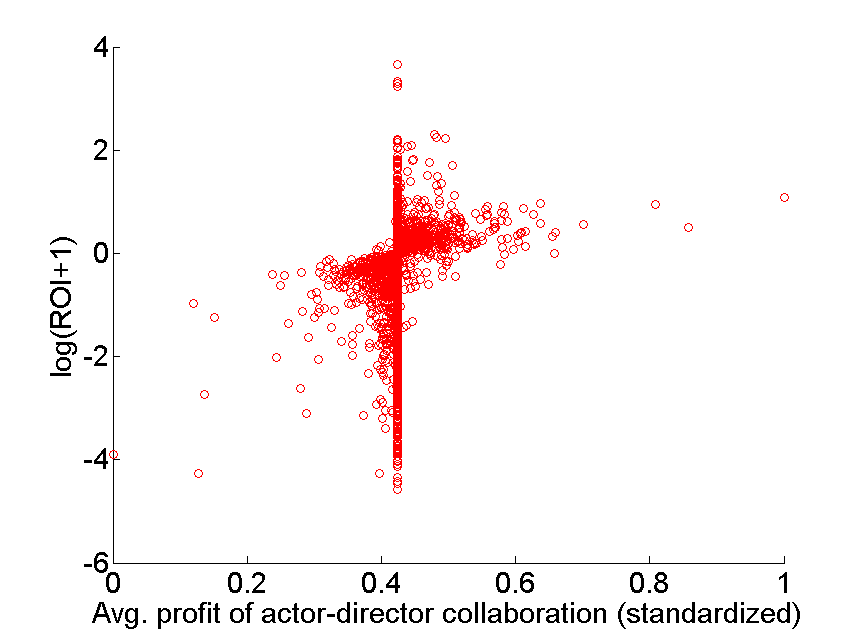}%
	\end{minipage}\protect\protect\protect\protect\protect\caption{The traditional measure of star power (total actor gross) vs. profit (left), and our top star power measure (avg. profit of past actor-director collaboration) vs. profit (right) \label{fig:star_power_ROI_compare}}
\end{figure}

\textcolor{rev}{By focusing on historical profitability records of both actors and directors, our results revealed some interesting findings about movie profitability. For example, director is an important factor for movies' profits. The top feature from our LASSO model is actually the average profit of previous actor-director collaboration. Distributions in Figure \ref{fig:star_power_ROI_compare} also show that the average profit of actor-director collaboration is a better indicator of log(ROI+1) than the traditional star power measure of total actor gross -- the rank correlation between the average profit of actor-director collaboration and log(ROI+1) is \textit{0.47}, while total actor gross has a rank correlation of \textit{0.29} with log(ROI+1). Also, having a star director is more indicative of profits than having a cast of star actors. Such findings actually contrasted with a few studies that even considered the effect of directors on movie success, albeit measured by box-office revenues. We conjectured that the difference may be due to measuring movie success using profits instead of revenues, and the usage of a larger dataset with movies whose success levels vary greatly. Although further investigations along this direction are beyond the scope of this research, we do believe that this is an interesting result that is worth exploring from team performance or marketing perspectives.}

\textcolor{rev}{Also, when it comes to predicting movie profits, actors' star power is better measured with their historical profits than with their gross. In fact, when we ranked actors by their total gross revenues and total profits, the ranking correlation is only moderate (Spearman coefficient of \textit{0.60}). Table \ref{tab:actor_ranking} lists top 10 actors by total revenues and total profits respectively, and there is only one (Julie Andrews) who appears on both lists. Although `big-name' movie stars are likely to attract quite a crowd, in the case of generating profits, the cost of casting such a star may not always be recouped via tickets sales. Thus when predicting movie profits, an actor's record in generating profits should be considered more than just his/her appearances in high-revenue movies.}

\begin{table}[htbp]
	\centering %
	\begin{tabular}{|c|c|c|}
		\hline 
		Rank & By Total Revenue & By Total Profit \tabularnewline
		\hline
		1 & Clark Gregg & Orlando Bloom\tabularnewline
		\hline 
		2 & Julie Andrews & Elijah Wood\tabularnewline
		\hline 
		3 & Dakota Fanning & Robert Pattinson \tabularnewline
		\hline 
		4 & Ashton Kutcher & Zoe Saldana \tabularnewline
		\hline 
		5 & Steve Carell & Mike Myers \tabularnewline
		\hline 
		6 & Morgan Freeman & Alan Rickman\tabularnewline
		\hline 
		7 &  Johnny Depp &  Julie Andrews \tabularnewline
		\hline 
		8 & Anna Kendrick & Samuel L. Jackson\tabularnewline
		\hline 
		9 & Bryce Dallas Howard & Gary Oldman\tabularnewline
		\hline 
		10 & Emma Roberts & Rupert Grint\tabularnewline
		\hline 
	\end{tabular}\protect\caption{Top 10 actors by total revenues and total profits.}
	\label{tab:actor_ranking}
\end{table}

\subsubsection{Teamwork and profitability}

\textcolor{rev}{From a teamwork perspective, we wanted to investigate the effect of expertise and diversity. We found that both can contribute to a movie's profits. Our new metric to measure how much expertise a cast has in a specific movie's genre -- Average Genre Expertise -- turns out to be positively related to profits (with a coefficient of \textit{0.007}). Along with the top positive coefficient for average actor-director collaboration profits, they have highlighted the importance of a cast's expertise and successful collaboration experience in the past. At the same time, diversity is also a positive predictor of profits. Among our diversity metrics, decrease in clustering coefficient, cast novelty, and the spanning of structural holes (measured by the decrease in the average shortest path lengths) all have positive coefficients in the LASSO model. In other words, a cast with members who have previous experience on a movie's genre, yet with some fresh faces, is beneficial for a movie's profits.}

\textcolor{rev}{In addition to showing the positive effect of team expertise and diversity on movies' profits, the results suggest that the manner, in which the expertise and diversity of a cast are measured, matters. Metrics used by previous research to capture experience and diversity -- average network centralities of a cast (both betweenness and degree centralities) -- have negative coefficients. This means that simply having a team of actors that are well connected in the collaboration network does not sufficiently make the cast experienced nor diverse. Instead, better measurement of expertise and diversity should focus on individuals' experience with different genres and the dynamic structure of the collaboration network, as our new features of Average Genre Expertise and the decrease in the average shortest path length did.}

\subsection{Limitations}

Admittedly, our study has limitations. For one, as we have mentioned,
there are possible biases that exist as far as our sampling goes.
As related by Figure \ref{fig:distrib_movies_genre}, we can see that
our dataset is relatively representative in terms of movie genres.
However, the limited availability of both budget and revenue data
may still introduce some bias to our dataset for the experiment, even though
we only excluded 17\% of all movies produced in the 11-year time period,
because of the lack of either kind of data.

Another limitation is that the profit we calculate is based on production
budget and box office revenue. For many movies, box office revenue
is only one of the sources of income. For example, Disney's animation
movies often gain a significant amount of their revenues from the
sale of movie merchandise, such as clothing and toys. Some movies
may also rely heavily on the sale and rental of DVDs besides ticket sales. However, capturing
these non-box-office revenues is more difficult as they may keep accumulating
many years after the release of a movie.


\section{Conclusions and Future Work}

In this study, we proposed the framework of a decision support system
(named MIAS) to aid investors' decisions on which movies to invest
in. MIAS learns from freely available historical data from various
sources and tries to predict the success of movies, which is defined
by profitability instead of revenue as most studies did. Also, because
investment decisions are made very early in the movie production process,
the features our system uses only leverages data that is available
early on. These features include both classic ones and novel ones that were proposed for the first time. They can be organized into four groups: ``who'' are on the cast from both individual and teamwork perspectives, ``what'' a movie
is about in terms of genre, rating, and topics of plot synopses, ``when''
a movie will be released, as well as ``hybrid'' features that match
``who'' with ``what'' and ``when'' with ``what'' features. 
Comparing with two benchmark methods, our experiments based on 11 years of movies showed that MIAS can do a better job in predicting the profitability of movies. \textcolor{rev}{In addition, to aid investors' decision on whether a proposed movie is worth investing, our system also allows them to conduct ``what-if'' analysis--different cast members, tweaks to the plot, or changes to the planned release time of a movie, can all be manipulated in order to experiment with what increases the chance of profitability. Besides movie investors, our system can also be helpful for other stakeholders in the movie industry who care about the possible financial success of a movie compared to the movie studio's expectations, such as cinemas that would like to decide whether to air a movie.}

Moreover, new features we proposed for this study were shown to make great contributions to 
the prediction of profitability. The framework of MIAS, as well as the ``what'', ``who'', ``when'', and ``hybrid'' features we extracted for
MIAS, can also be applied to other areas beyond movies. For example, it can be adapted
to predict the success of other creative works, which often requires
a team of contributors, whose content can be described with texts,
and for which timing is important, such as research papers, grant
proposals, operas, etc. 

\textcolor{rev}{Although the focus of this study is on the design of a decision support system, its outcomes could potentially have theoretical implications as well. Our regression analysis revealed the effects of key factors of movie profitability. Some findings are different from previous studies. For example, our study highlighted the importance of directors for movie profitability. Our new methods of quantifying factors suggested by past theoratical studies (e.g., actor star powers, team expertise, and team diversity) also helps to better capture them and worked very well in the context of profit prediction. We hope these findings will inspire future theoretical research in areas such as marketing, creative works, and team performance.}

There are also several directions for future research. For example, as we have matched ``what'' with ``when'' and ``what'' with
``who'', it would be interesting to match ``who'' with ``when'' to capture whether the popularity or an actor or director is on the
rise or decline. This will require more fine-grained analysis of an actor or director's career. Also, while we have considered the formal
collaboration network among actors, the informal friendship network
may also play a role in the formation and success of a cast, although
such a network will be more difficult to capture. \textcolor{rev}{Additionally, the popularity of movies and the rise of the internet has brought with it a new community of fans, including those willing to document entire scripts. These fan-based endeavors are published on sites such as the Internet Movie Script Database. Another interesting future direction for research would be to collect full-length scripts of a large number of movies and to then analyze the scripts, instead of the plot synopses. With movie scripts we can get more fine-grained topic distribution vectors, as well as many other novel features, such as script cadence. We also intend on adding more features to our model, including those that more definitively speak to consumer spending power, such as external economic indices, as well as those that take into account the types of movies that are most suited to certain times of the year (i.e. is it best to release Christmas-themed movies at Christmas time?).}

\singlespacing 

\bibliography{dss_bib}

\newpage{}
\section{Appendix}

\begin{table}[h]
	\centering %
	\begin{tabular}{|l|c|c|c|c|c|}
		\hline 
		Measure  & \multicolumn{1}{l|}{J48} & \multicolumn{1}{l|}{naive Bayes} & \multicolumn{1}{l|}{Multilayer Perceptron} & \multicolumn{1}{l|}{SVM} & \multicolumn{1}{l|}{Log. Regression}\tabularnewline
		\hline 
		AUC  & 0.752  & 0.704 & 0.698 & 0.643 & 0.831 \tabularnewline
		\hline 
		Accuracy  & 0.796  & 0.718 & 0.718 & 0.771 & 0.809 \tabularnewline
		\hline 
		Precision (Pos. Class)  & 0.673  & 0.539 & 0.548 & 0.807 & 0.785  \tabularnewline
		\hline 
		Recall (Pos. Class)  & 0.631  & 0.461 & 0.386 & 0.32 & 0.507 \tabularnewline
		\hline 
		Precision (Neg. Class)  & 0.845  & 0.781 & 0.764 & 0.767 & 0.815 \tabularnewline
		\hline 
		Recall (Neg. Class)  & 0.867  & 0.804 & 0.862 & 0.967 & 0.94 \tabularnewline
		\hline 
	\end{tabular}\protect\caption{Additional binary classification results with top 30\% ROIs as the decision boundary.}
	\label{tab:append_30_roi} 
\end{table}

\begin{table}[h]
	\centering %
	\begin{tabular}{|l|c|c|c|c|c|}
		\hline 
		Measure  & \multicolumn{1}{l|}{J48} & \multicolumn{1}{l|}{naive Bayes} & \multicolumn{1}{l|}{Multilayer Perceptron} & \multicolumn{1}{l|}{SVM} & \multicolumn{1}{l|}{Log. Regression}\tabularnewline
		\hline 
		AUC  & 0.807  & 0.785 & 0.775 & 0.692 & 0.908 \tabularnewline
		\hline 
		Accuracy  & 0.876  & 0.782 & 0.813 & 0.856 & 0.884 \tabularnewline
		\hline 
		Precision (Pos. Class)  & 0.718  & 0.491 & 0.598 & 0.832 & 0.797  \tabularnewline
		\hline 
		Recall (Pos. Class)  & 0.694  & 0.543 & 0.394 & 0.407 & 0.616 \tabularnewline
		\hline 
		Precision (Neg. Class)  & 0.918  & 0.872 & 0.849 & 0.858 & 0.902 \tabularnewline
		\hline 
		Recall (Neg. Class)  & 0.926  & 0.847 & 0.928 & 0.978 & 0.957 \tabularnewline
		\hline 
	\end{tabular}\protect\caption{Additional binary classification  results with $ROI\geq67\%$ as the decision boundary.}
	\label{tab:append_quart_st_dev_profit} 
\end{table}

\begin{table}[h]
	\centering %
	\begin{tabular}{|l|c|c|c|c|c|c|}
		\hline 
		Measure  & \multicolumn{1}{l|}{J48} & \multicolumn{1}{l|}{naive Bayes} & \multicolumn{1}{l|}{MLP} & \multicolumn{1}{l|}{SVM} & \multicolumn{1}{l|}{Log. Reg.} & \multicolumn{1}{l|}{LogitBoost}\tabularnewline
		\hline 
		AUC  & 0.733  & 0.527 & 0.687 & 0.714 & 0.806 & 0.813 \tabularnewline
		\hline 
		Accuracy  & 0.655  & 0.539 & 0.515 & 0.584 & 0.64 & 0.652 \tabularnewline
		\hline 
		Precision (Pos. Class)  & 0.724 & 0.612 & 0.538 & 0.716 & 0.491 & 0.78  \tabularnewline
		\hline 
		Recall (Pos. Class)  & 0.733  & 0.451 & 0.547 & 0.58 & 0.665 & 0.628 \tabularnewline
		\hline 
		Cost  & 1095  & 1433 & 1648 & 1343 & 1089 & 1041 \tabularnewline
		\hline 
	\end{tabular}\protect\caption{Additional multi-class prediction results for the first decision boundary.}
	\label{tab:append_multiclass_first_board} 
\end{table}

\begin{table}[h]
	\centering %
	\begin{tabular}{|l|c|c|c|c|c|c|}
		\hline 
		Measure  & \multicolumn{1}{l|}{J48} & \multicolumn{1}{l|}{naive Bayes} & \multicolumn{1}{l|}{MLP} & \multicolumn{1}{l|}{SVM} & \multicolumn{1}{l|}{Log. Reg.} & \multicolumn{1}{l|}{LogitBoost}\tabularnewline
		\hline 
		AUC  & 0.731  & 0.747 & 0.693 & 0.722 & 0.819 & 0.826 \tabularnewline
		\hline 
		Accuracy  & 0.678  & 0.626 & 0.577 & 0.648 & 0.697 & 0.716 \tabularnewline
		\hline 
		Precision (Pos. Class)  & 0.708 & 0.581 & 0.475 & 0.688 & 0.764 & 0.84  \tabularnewline
		\hline 
		Recall (Pos. Class)  & 0.732  & 0.498 & 0.571 & 0.523 & 0.651 & 0.619 \tabularnewline
		\hline 
		Cost  & 933  & 1120 & 1059 & 967 & 816 & 759 \tabularnewline
		\hline 
	\end{tabular}\protect\caption{Additional multi-class prediction results for the second decision boundary.}
	\label{tab:append_multiclass_second_board} 
\end{table}

\end{document}